\documentclass{article}

\PassOptionsToPackage{numbers}{natbib}

     \usepackage[preprint]{neurips_2020}

\usepackage[utf8]{inputenc} 
\usepackage[T1]{fontenc}    
\usepackage{hyperref}       
\usepackage{url}            
\usepackage{booktabs}       
\usepackage{amsfonts}       
\usepackage{nicefrac}       
\usepackage{microtype}      

\title{Probabilistic Optimal Transport \\ based on Collective Graphical Models}

\usepackage{amssymb,bm,ascmac,amsthm,multirow,bigdelim, algorithmic, comment, mathtools}

\usepackage[linesnumbered,ruled,vlined]{algorithm2e}

\newcommand{\ba}{\bm{a}}
\newcommand{\bb}{\bm{b}}
\newcommand{\bc}{\bm{c}}

\newcommand{\bu}{\bm{u}}
\newcommand{\bv}{\bm{v}}
\newcommand{\bt}{\bm{t}}

\newcommand{\bx}{\bm{x}}
\newcommand{\by}{\bm{y}}
\newcommand{\bz}{\bm{z}}
\newcommand{\bw}{\bm{w}}

\newcommand{\bX}{\bm{X}}

\newcommand{\bpi}{\bm{\pi}}

\newcommand{\dI}{\mathbb{I}}
\newcommand{\dL}{\mathbb{L}}
\newcommand{\dZ}{\mathbb{Z}}
\newcommand{\dZp}{\mathbb{Z}_{\geq 0}}
\newcommand{\dR}{\mathbb{R}}
\newcommand{\dRp}{\mathbb{R}_{\geq 0}}
\newcommand{\defeq}{:=}
\newcommand{\seq}[2]{#1 \in [#2]}

\newcommand{\bone}[1]{\bm{1}_{#1}}
\newcommand{\bzero}[1]{\bm{0}_{#1}}
\newcommand{\ta}{\bar{a}}
\newcommand{\tb}{\bar{b}}
\newcommand{\tx}{\bar{x}}
\newcommand{\bta}{\bm{\ta}}
\newcommand{\btb}{\bm{\tb}}
\newcommand{\btx}{\bm{\tx}}
\newcommand{\TP}{T}
\newcommand{\mTP}{\TP}
\newcommand{\KL}{\mathrm{KL}}
\newcommand{\gKL}{\widetilde{\mathrm{KL}}}

\newcommand{\argmin}{\mathop{\rm arg~min}\limits}
\newcommand{\KLprox}[2]{\mathrm{Prox}_{#1}^{\mathrm{KL}}\left( #2 \right)}
\newcommand{\UU}[2]{U(#1, #2)}
\newcommand{\UF}[2]{U^F(#1, #2)}

\newtheorem*{theorem*}{Theorem}
\newtheorem{proposition}{Proposition}
\newtheorem*{proposition*}{Proposition}

\newtheorem*{definition*}{Definition}

\newtheorem*{corollary*}{Corollary}

\newtheorem*{lemma*}{Lemma}

\author{
  Yasunori Akagi \\
  NTT Service Evolution Laboratories\\
  \texttt{yasunori.akagi.cu@hco.ntt.co.jp} \\
  \And
  Yusuke Tanaka \\
  NTT Service Evolution Laboratories \\
  \texttt{yusuke.tanaka.rh@hco.ntt.co.jp} \\
  \AND
  Tomoharu Iwata \\
  NTT Communication Science Laboratories \\
  \texttt{tomoharu.iwata.gy@hco.ntt.co.jp} \\
  \And
  Takeshi Kurashima \\
  NTT Service Evolution Laboratories \\
  \texttt{takeshi.kurashima.uf@hco.ntt.co.jp} \\
  \And
  Hiroyuki Toda \\
  NTT Service Evolution Laboratories \\
  \texttt{hiroyuki.toda.xb@hco.ntt.co.jp} \\
}

\begin{document}

\maketitle

\begin{abstract}
  Optimal Transport (OT) is being widely used in various fields such as machine learning and computer vision, as it is a powerful tool for measuring the similarity between probability distributions and histograms.
  In previous studies, OT has been defined as the minimum cost to transport probability mass from one probability distribution to another.
  In this study, we propose a new framework in which OT is considered as a maximum a posteriori (MAP) solution of a probabilistic generative model.
  With the proposed framework, we show that OT with entropic regularization is equivalent to maximizing a posterior probability of a probabilistic model called Collective Graphical Model (CGM), which describes aggregated statistics of multiple samples generated from a graphical model.
  Interpreting OT as a MAP solution of a CGM has the following two advantages: (i) We can calculate the discrepancy between noisy histograms by modeling noise distributions. Since various distributions can be used for noise modeling, it is possible to select the noise distribution flexibly to suit the situation. (ii) We can construct a new method for interpolation between histograms, which is an important application of OT. The proposed method allows for intuitive modeling based on the probabilistic interpretations, and a simple and efficient estimation algorithm is available.  
  Experiments using synthetic and real-world spatio-temporal population datasets show the effectiveness of the proposed interpolation method.

\end{abstract}

\section{Introduction}
Optimal Transport (OT) is a framework for measuring the similarity between probability distributions or histogram data.
It has been applied to various major machine learning fields such as classification \cite{Frogner2015}, transfer learning \cite{Courty2017}, and generative modeling \cite{Arjovsky2017}, and its effectiveness has been confirmed. 
Furthermore, the geometric structure of the histogram space introduced by OT makes it possible to perform important operations on histograms such as interpolation or determination of a representative point between multiple histogram data. 
These OT-based operations have increased the importance of OT in fields such as computer graphics \cite{Solomon2015}\cite{Peyre2019} or spatio-temporal data mining \cite{Solomon2014}\cite{Roberts2017}. 

Basic OT is defined as the minimum cost to transport probability mass from one probability distribution to another. 
Although many variants of OT have also been proposed and used such as Sinkhorn distance \cite{Cuturi2013}, these are also defined as the minimum transport cost of probability mass with special regularization terms.

In this paper, we present a new definition of OT, where it is defined by a maximum a posteriori (MAP) solution of a probabilistic generative model, 
and develop a new methodology for interpolation based on this definition.
To do this, we utilize Collective Graphical Model (CGM) \cite{Sheldon2011}, which is a probabilistic generative model for describing aggregated statistics of multiple samples generated from a graphical model.  
We show that the objective function of OT with entropic regularization can be written as the approximated negative log of the joint distribution of aggregated statistics and observation in a certain CGM. 
Using this fact, it can be shown that OT with entropic regularization is equivalent to MAP inference of the CGM under observation. 

Interpreting and formulating OT as a MAP solution of a probabilistic model has several advantages. 
First, even if the data cannot be accurately observed, we can calculate OT by probabilistically modeling the effect of the noise. 
Since various distributions can be used for noise modeling, it is possible to select the distribution of noise flexibly to suit the situation.
For some noise distribution settings, the objective function is the same as that of the existing unbalanced OT \cite{Benamou2003}.

Second, we can construct a new method based on probabilistic modeling for interpolation between histograms, which is an important application of OT. 
In the proposed method, we can easily design potentials of the underlying graphical model, thus the transport processes can be controlled so that interpolation result is intuitive. 
We derive simple and efficient estimation algorithms based on message passing on CGM. 
In addition, the proposed method can be generalized naturally to the histogram propagation problem on general trees, just as Wasserstein barycenter \cite{Agueh2011} can be generalized to Wasserstein propagation \cite{Solomon2014}. 

Recently, Singh et al. \cite{Singh2020} pointed out the relationship between OT and CGM: they propose a new inference algorithm for CGM based on Sinkhorn-Knopp algorithm, which is often used in OT studies. 
Their work focused on just a specific task, i.e., contingency table estimation,  which has been addressed previously in CGM studies. On the other hand, we aim to provide a new formulation of OT as a MAP inference of CGM, which allows us to design OT via a probabilistic perspective. This contribution has a significant potential impact because it can be applied to various tasks which have been addressed in OT studies; for example, our formulation can be used for effectively solving a histogram interpolation task described in Section \ref{sec: histogram interpolation}.
\section{Backgrounds}
\subsection{Optimal Transport}
Optimal transport (OT) is a theory about how probabilistic mass can be transported from one probabilistic distribution to another. 
The minimum transportation cost, called OT distance, can be used as a metric that quantifies the distance between two probability distributions. 
OT distance has recently been shown to offer better performance than traditional distance measures between probability distributions, such as KL divergence and total variation distance, and is increasingly being used in various fields of machine learning \cite{Courty2017}\cite{Arjovsky2017}\cite{Frogner2015}.

We explain here the mathematical formulation of OT in a discrete state space. 
Let $[n] \defeq \{1, \ldots, n\}$ and $\Sigma_n^F \defeq \left\{ \bm{a} \in \mathbb{R}_{\geq 0}^n \mid \sum_{\seq{i}{n}} a_i = F\right\}$ is the set of $n$-dimensional non-negative vectors with total sum $F$. 
Of particular note, $\Sigma_n^1$ is the set of $n$-dimensional probabilistic vectors. 
For $\ba, \bb \in \Sigma_n^F$, we define transportation polytope $\UF{\ba}{\bb} \defeq \left\{ \mTP \in \mathbb{R}_{\geq 0}^{n \times n} \mid \mTP \bone{n} = \ba, \mTP^\top \bone{n} \defeq \bb \right\}$, where $\bone{n} = [1, \ldots, 1]^\top \in \dR^n$.  
Then, the OT distance between $\ba$ and $\bb$ with cost matrix $C \in \mathbb{R}^{n \times n}$ is defined by 
$D_C(\ba, \bb) \defeq \mathrm{min}_{\mTP \in \UF{\ba}{\bb}} \mathcal{G}_C(\TP)$, where $\mathcal{G}_C(\TP)$ is the cost of the transportation matrix $\mTP$ ($\mathcal{G}_C(\TP) \defeq \sum_{i, j} C_{ij} \TP_{ij}$). 
Although the optimization problem in the definition can be solved in polynomial time via linear programming, computation cost becomes excessive when $n$ is large since its time complexity is $O(n^3)$ \cite{Peyre2019}. 

In order to avoid excessive computation costs, a variant of OT distance, called Sinkhorn distance, was proposed \cite{Cuturi2013}. 
Sinkhorn distance is defined as the optimum value of the optimization problem 
$D_C^{\epsilon}(\ba, \bb) \defeq \mathrm{min}_{\mTP \in \UF{\ba}{\bb}} \mathcal{G}_C^{\epsilon}(\TP)$, where $\mathcal{G}_C^{\epsilon}(\TP) \defeq \sum_{i, j} \left[ C_{ij} \TP_{ij} + \epsilon \  \TP_{ij} \log \TP_{ij} \right]$ and $\epsilon \in \dR_{>0}$ is a hyperparameter. 
The difference from the original OT distance is the term $\epsilon \ \TP_{ij} \log \TP_{ij} $, which is the negative entropy of transportation matrix $T$. 
Sinkhorn distance can be calculated efficiently by the Sinkhorn-Knopp algorithm \cite{Knight2008}, which consists of iterative matrix multiplication.   
\subsection{Collective Graphical Model}
Collective Graphical Model (CGM) is a probabilistic generative model that describes the characteristics of aggregated statistics of multiple samples drawn from a certain graphical model \cite{Sheldon2011}. 
Let $G=(V, E)$ be an undirected tree graph (i.e., contains no cycles). We consider a pairwise graphical model over discrete random variable $\bX \defeq (X_1, \ldots, X_{|V|})$ defined by
$
  \Pr(\bX=\bx) = (1/Z) \prod_{(u, v) \in E} \phi_{uv}(x_u, x_v), 
$
where $\phi_{uv}(x_u, x_v)$ is a local potential function on edge $(u, v)$ and $Z \defeq \sum_{\bx} \prod_{(u, v) \in E} \phi_{uv}(x_u, x_v)$ is a partition function for normalization. 
In this paper, we assume that $x_u$ takes values on the set $[n]$ for all $u \in V$. 

We draw ordered samples $\bX^{(1)}, \ldots, \bX^{(F)}$ independently from the graphical model.  
We define node contingency table $\bt_u = (t_u(x_u): x_u \in [n])$ for node $u$ and edge contingency table $\bt_{uv} = (t_{uv}(x_u, x_v): x_u, x_v \in [n])$ for edge $(u, v)$, which are the vectors whose entries are the number of occurrences of particular variable settings: 
\begin{align*}
  t_u(x_u) \defeq \left| \{ f \in [F] \mid  X_u^{(f)}=x_u \} \right|, \  t_{uv}(x_{u}, x_{v}) \defeq \left| \{ f \in [F] \mid  X_u^{(f)}=x_u \land X_v^{(f)}=x_v \} \right|, 
\end{align*}
where $t_u(x_u)$ is the number of samples which satisfies $X^{(f)}_{u} = x_u$, and $t_{uv}(x_u, x_v)$ is the number of samples which satisfies $X_u^{(f)}=x_u$ and $X_v^{(f)}=x_v$. 
In CGM, whole observation $\by:= \{\by_u\}_{u \in V} \cup \{\by_{u, v}\}_{(u, v) \in E}$ is generated by adding noise to all contingency tables $\bt :=\{\bt_u\}_{u \in V} \cup \{\bt_{u, v}\}_{(u, v) \in E}$. 
In this paper, we assume that only node observation $\{ \by_u \}_{u \in V}$ is given, but edge observation $\{ \by_{uv} \}_{(u, v) \in E}$ is not.

We address the problem of inferring contingency table $\bt$ from observation $\by$. 
In order to solve this, we try to get $\bt^*$, which maximizes posterior probability $\Pr(\bt|\by)$. 
This approach is called MAP inference in CGM. 
MAP inference is one of the main topics of CGM studies \cite{Sheldon2013a}\cite{Sun2015}\cite{Nguyen2016}, because it is important for various CGM-based tasks, such as parameter estimation of the graphical model. 

Since $\Pr(\bt|\by) = \Pr(\bt, \by)/\Pr(\by)$ from Bayes' rule, it is sufficient to maximize the joint probability $\Pr(\bt, \by) = \Pr(\bt) \Pr(\by | \bt)$, where $\Pr(\by | \bt)$ is the noise distribution associated with observation. $\Pr(\bt)$ is called CGM distribution and calculated as follows \cite{Sun2015}: 
\footnotesize
\begin{align}
  \Pr(\bt) = \frac{F!}{Z^F} \cdot \frac{\prod_{u \in V} \prod_{x_u \in [n]} \left(t_u(x_u)! \right)^{\nu_u - 1}}{\prod_{(u, v) \in E} \prod_{x_u, x_v \in [n]} t_{uv}(x_u, x_v)!}
  \cdot \prod_{(u, v) \in E} \prod_{x_u, x_v \in [n]} \phi(x_u, x_v)^{t_{uv}(x_u, x_v)} 
  \cdot \dI (\bt \in \dL_F^{\dZ}), \\
  \dL_F^\dZ \defeq
  \left\{
  \bt \in \dZp^{|\bt|} \middle| F = \sum_{x_u \in [n]} t_u(x_u) \  \forall u \in V, t_u(x_u) = \sum_{x_v \in [n]} t_{uv}(x_u, x_v) \ \forall (u, v) \in E, x_u \in [n] \right\}. 
\end{align}
\normalsize
Here, $\dL_F^\dZ$ is the set of contingency tables $\bt$ that satisfy the consistency of counts among the number of samples $F$, node contingency tables $\bt_u$, and edge contingency tables $\bt_{uv}$. 
Let $\dL_F^\dR$ be the set obtained by removing integrality constraints from $\dL_F^\dZ$. 
Although exact MAP inference is known to be intractable \cite{Sheldon2013a}, by relaxing the integrality constraints (i.e. replacing the feasible set $\dL_F^\dZ$ with $\dL_F^\dR$), taking the negative log of the objective function, and applying Stirling's approximation, we get a tractable approximate MAP inference problem: 
\footnotesize
\begin{align}
  &\min_{\bz \in \dL_F^\dR} \mathcal{L}(\bz) = E_{\rm CGM}(\bz) - H_{\rm B}(\bz) - F \log F + F \log Z, \label{formula: approx CGM} \\
  &E_{\rm CGM}(\bz) \defeq - \sum_{(u, v) \in E} \sum_{x_u, x_v \in [n]} z_{uv}(x_u, x_v) \log \phi_{uv}(x_u, x_v) - \log \Pr(\by|\bz), \\
  &H_{\rm B}(\bz) \defeq - \sum_{(u, v) \in E} \sum_{x_u, x_v \in [n]} z_{uv}(x_u, x_v) \log z_{uv}(x_u, x_v) + \sum_{u \in V} (\nu_u-1)\sum_{x_u \in [n]} z_u(x_u) \log z_u(x_u). 
\end{align}
\normalsize
where $\mathcal{L}(\bz)$ is the approximated negative log joint probability $\mathcal{L}(\bz) \approx- \log \Pr(\bz, \by)$. 
Note that integer-valued variable $\bt$ is replaced by real-valued variable $\bz$ via continuous relaxation. 
This approximation is often used in CGM studies\cite{Sheldon2013a}\cite{Sun2015}\cite{Nguyen2016}. 
This is a convex programming problem and known to be efficiently solved by message-passing style algorithms \cite{Sun2015}. 

\section{Relationship between OT and CGM} \label{sec: OT vs CGM}
In this section, we newly define OT based on CGM and detail the relationship between OT and CGM. 
We consider a CGM on $P_2$, where $P_2$ is a path graph with two nodes $\{1, 2\}$. 
In this case, the contingency table and observation are generated as follows: (i) $F$ samples $\{ (X_1^{(f)}, X_2^{(f)})\}_{f=1}^F$ are drawn from the graphical model on $P_2$ with potential $\phi_{1, 2}(i, j)$. (ii) Contingency tables are determined by aggregating and counting the sample values: $t_1(i) = \left| \{f \mid X_1^{(f)} = i \} \right|$, $t_2(i) = \left| \{f \mid X_2^{(f)} = i \} \right|$, $t_{1, 2}(i, j) = \left| \{f \mid X_1^{(f)} = i \land X_2^{(f)} = j \} \right|$. (iii) Observations $\by_1$ and $\by_2$ are generated by adding noise to $\bt_1$ and $\bt_2$. 
We here consider here noiseless observations, i.e., $\by_1 = \bt_1$ and $\by_2 = \bt_2$, while noisy observations are considered in Section {\ref{sec: noise}}
We write $\psi_{ij} \defeq \phi_{1, 2}(i, j), \TP_{ij} \defeq z_{1, 2}(i, j), a_i \defeq y_1(i), b_i \defeq y_2(i)$ for simplicity. 
\begin{proposition} \label{prop: objective}
  For all $T \in \UF{\ba}{\bb}$, 
  $
  \mathcal{G}_C^1(\tau) = \mathcal{L}(\TP)/F - \log Z
  $
  , where $\tau_{ij} \defeq \TP_{ij}, C_{ij} \defeq - \log \psi_{ij}$. 
\end{proposition}
All proofs are given in the Appendix. 
Proposition \ref{prop: objective} states that the approximated negative log of joint probability $\mathcal{L}(T)$ can be expressed as transportation cost of the corresponding OT instance with entropic regularization ($\epsilon=1$). 
Based on Proposition \ref{prop: objective}, we reveal the relationship between Sinkhorn distance and MAP inference of this CGM. 

\begin{proposition} \label{proposition: OT vs CGM}
  Let $\bm{\alpha} \defeq \ba/F,\bm{\beta} \defeq \bb/F$. 
  For all $\ba, \bb \in \Sigma_n^F$, 
  \begin{align}
  \mathcal{D}_C^1(\bm{\alpha}, \bm{\beta}) = \frac{1}{F} \min_{T \in \UU{\ba}{\bb}} \left[ \mathcal{L}(T) \right] - \log Z \label{formula: OT vs CGM}. 
  \end{align}
\end{proposition}
Proposition \ref{proposition: OT vs CGM} says that Sinkhorn distance with $\epsilon=1$ can be described by the approximated maximum joint probability of the CGM. 
Moreover, because $\Pr(\mTP|\ba, \bb) \propto \Pr(\mTP, \ba, \bb)\approx \mathcal{L}(T)$, it can be seen that the MAP inference in the CGM given observation $\ba$ and $\bb$ is equivalent to calculating Sinkhorn distance with $\epsilon=1$.
The RHS of (\ref{formula: OT vs CGM}) can be interpreted as the average value of approximated negative log-likelihood of the CGM per one sample.  
Since Stirling's approximation becomes precise when $F$ is sufficiently large, the RHS of (\ref{formula: OT vs CGM}) approaches to the exact average of negative log-likelihood when $F  \to \infty$. 
Thus, formula (\ref{formula: OT vs CGM}) states that the exact average of negative log-likelihood per one sample of the CGM equals to Sinkhorn distance with $\epsilon=1$ when $F \to \infty$. 

This relationship yields several insights: 
(i) OT distance has often been thought of as the cost of transportation, but it can be reinterpreted as the maximum value of the joint (or posterior) probability of a certain probabilistic generative model. 
(ii) The hyperparameter setting $\epsilon=1$ has a special meaning based on probabilistic interpretation. This fact can be a useful clue in determining this hyperparameter value. 
(iii) Probabilistic interpretation allows us to extend OT via probabilistic modeling. For example, we can consider OT with noisy observations (Section \ref{sec: noise}), or construct a interpolation method between histograms based on probabilistic modeling (Section \ref{sec: histogram interpolation}). 

\section{OT with noisy observations} \label{sec: noise}
One advantage of taking the probabilistic interpretation of OT is that we can calculate the discrepancy between noisy histograms by modeling noise distributions. 
We consider a CGM on graph $P_2$ and use the same notations except for  $\ta_i \defeq z_1(i), \tb_i \defeq z_2(i)$. 
The observations $\ba, \bb$ are assumed to be generated according to distributions $\Pr(\ba | \bta), \Pr(\bb | \btb)$, which represent observation noise. 
In this case, from (\ref{formula: approx CGM}), we have 
\footnotesize
\begin{align}
  \frac{1}{F} \min_{T \in \dR_{\geq 0}^{n \times n}} \mathcal{L}(T) - \log Z
  &= \min_{\tau \in \dR_{\geq 0}^{n \times n}} \left[ \mathcal{G}_C^1(\tau) - \frac{1}{F} \log \Pr(F \bm{\alpha} \mid F \tau \bone{n}) - \frac{1}{F} \log \Pr(F \bm{\beta} \mid F \tau^\top \bone{n}) \right] \label{formula: noisy OT}. 
\end{align}
\normalsize
Based on the similarity between LHS of (\ref{formula: noisy OT}) and RHS of (\ref{formula: OT vs CGM}), we define OT with noisy observations by the RHS of (\ref{formula: noisy OT}). 
The difference from noiseless OT is the second and the third term: those terms represent discrepancy between observed histograms and marginals of the transportation matrix. 

We can utilize various probabilistic distributions for $\Pr(\ba \mid \bta), \Pr(\bb \mid \btb)$. 
For example, when we use i.i.d. Gaussian distributions
$
a_i \sim \mathcal{N}(\ta_i, F \sigma^2), 
b_i \sim \mathcal{N}(\tb_i, F \sigma^2), 
$
the RHS of (\ref{formula: noisy OT}) asymptotically becomes 
$
  \min_{\tau} \left[ \mathcal{G}_C^1(\tau) + (1/2 \sigma^2) \| \bm{\alpha} - \tau \bone{n} \|_2^2 + (1/2 \sigma^2) \| \bm{\beta} - \tau^\top \bone{n} \|_2^2 \right], 
$
when $F \to \infty$. 
When we use i.i.d. Poisson distributions
$ 
a_i \sim \mathrm{Poisson}(\ta_i), 
b_i \sim \mathrm{Poisson}(\tb_i), 
$
the RHS of (\ref{formula: noisy OT}) becomes
  $
  \min_{\tau} \left[ \mathcal{G}_C^1(\tau) +\gKL(\bm{\alpha} \| \tau \bone{n}) + \gKL(\bm{\beta} \| \tau^\top \bone{n}) \right], 
  $
by applying Stirling's approximation to log factorial, where $\gKL$ is the generalized KL divergence 
$
  \gKL(\bm{w} \| \bm{z}) = \sum_{\seq{i}{n}} w_i \log \left( {w_i}/{z_i} \right) - \sum_{\seq{i}{n}} w_i + \sum_{\seq{i}{n}} z_i
$. 

These formulations are closely related to unbalanced OT \cite{Benamou2003}.
Unbalanced OT is a method to measure the discrepancy between two histograms that have different total mass. 
In unbalanced OT, the differences between histograms and marginals of transportation matrix are added to the objective function as a penalty term. 
The Gaussian noise case agrees with unbalanced OT with squared 2-norm regularization\cite{Benamou2003}\cite{Blondel2018}. 
The Poisson noise case is similar to the relaxed OT in \cite{Frogner2015} (note that $\gKL(\tau \bone{n} \| \bm{\alpha}) + \gKL(\tau^\top \bone{n} \| \bm{\beta})$ is used in \cite{Frogner2015}, which is a bit different from ours). 
Thus, unbalanced OT can also be interpreted as the maximum value of the negative log of joint probability in a CGM. 
This relationship gives us a clue to select the appropriate penalty functions for unbalanced OT: we can measure the discrepancy between histograms appropriately by choosing the penalty terms derived from the noise distribution present in the situation of interest. 

We can solve the optimization problem in the RHS of (\ref{formula: noisy OT}) by the generalized Sinkhorn algorithm using KL proximal operator, when $- \log \Pr(\bx|\btx)$ is a convex function with respect to $\btx$  \cite{Chizat2018}. For more details, please see the Appendix. 

\section{Probabilistic interpolation between histograms} \label{sec: histogram interpolation}
In this section, we propose a new method for interpolating between histograms, which is an important application of OT, based on the probabilistic interpretation of OT. 
Here we propose two methods: the first ones is a naive method using undirected graphical model, and the second one is an advanced method via continuous time Markov chain for resolving shortcomings of the first one. 

\begin{figure}[t]
  \begin{center}
  \includegraphics[width=130mm]{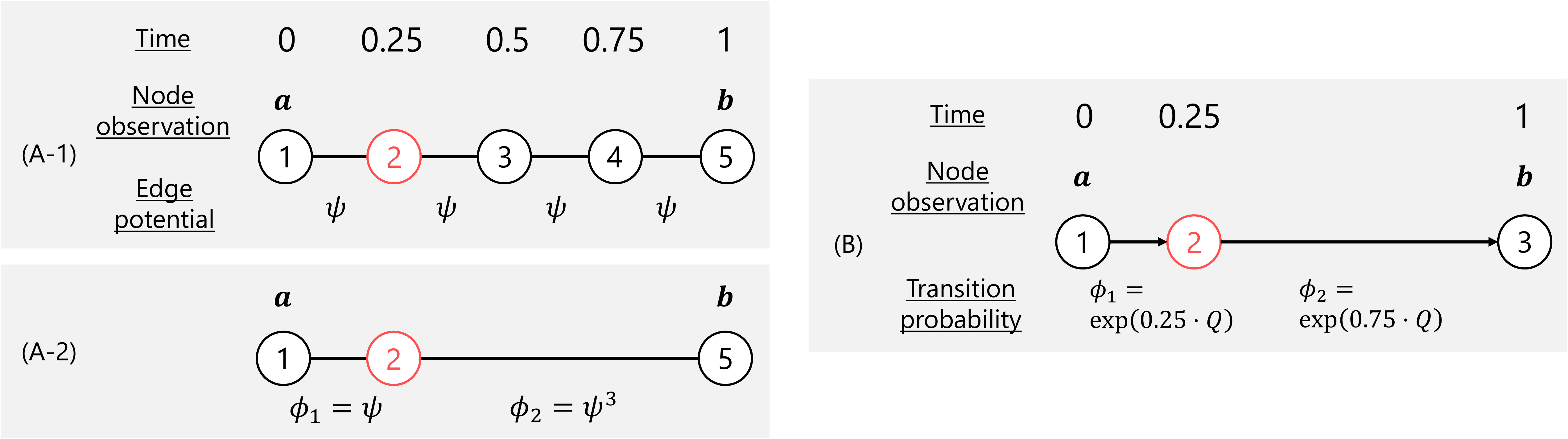}
  \end{center}
  \caption{Examples of graphical model in interpolating the histogram at $t=0.25$. In both cases, interpolation is conducted by estimating the node contingency table at the node corresponding to $t=0.25$ (red). In (A-1) and (A-2), $N=5$ and $k=2$. } 
  \label{fig: interpolation_graphcial_model}
\end{figure}

\subsection{Problem settings and previous methods}
We consider the following problem setting.
We are given histogram data $\ba$ at time $0$ and the histogram data $\bb$ at time $1$. 
Our task is to estimate the histogram at time $t\ (0 < t < 1)$.
This kind of interpolation problem has become one of the major applications of OT \cite{Solomon2014} \cite{Solomon2015}. 
In previous work, the histogram at time $t$ is given by the optimum solution of 
\begin{align}
  \argmin_{\bc \in \Sigma_n^F } \left[ (1-t) \cdot \mathcal{D}^{\epsilon}_{C}(\ba, \bc) + t \cdot \mathcal{D}^{\epsilon}_{C}(\bb, \bc) \right].  \label{formula: WB}
\end{align}
This result can be considered as the Fr\'echet mean in the metric space introduced by OT. 
When we use Euclidean distance as cost function $C$ and $\epsilon=0$, the estimated result is called Wasserstein barycenter of the two histograms $\ba, \bb$ \cite{Agueh2011}. 

\subsection{Undirected graphical model-based method} \label{subsec: undirected graphical model}
We consider an undirected graphical model on path graph $P_N$ ((A-1) in Figure \ref{fig: interpolation_graphcial_model}). 
$N$ is an integer such that $(k-1)/(N-1) \approx t$ holds for some integer $k \in \{2, \ldots, N-1\}$. 
Suppose that potentials of edges $\phi_{u, u+1}$ are common to all edges $(u, u+1)$ and can be written as  $\phi_{u, u+1}(i, j) = \psi_{ij}$  using some $\psi \in \dR^{n\times n}$.
In the proposed method, we obtain an interpolation between histograms $\ba$ and $\bb$ by solving a MAP inference problem $\min_{\bz \in \dL_F^\dR} -L(\bz)$ of the CGM on this graphical model, when the observation at node 1 is $\ba$ and at node $N$ is $\bb$. 
The result of interpolation is given by $\bz^*_{k}$, where $\bz^*$ is the MAP solution. 
This means that the estimated histogram is the contingency table with maximum posterior probability at node $k$.

\newcommand{\potl}{\phi_1}
\newcommand{\potr}{\phi_2}
\newcommand{\tpl}{T_1}
\newcommand{\tpr}{T_2}
\newcommand{\potlij}{\phi_{1ij}}
\newcommand{\potrij}{\phi_{2ij}}
\newcommand{\tplij}{T_{1ij}}
\newcommand{\tprij}{T_{2ij}}

When we need only the histogram at node $k$, it is sufficient to consider an CGM on path graph $P_3$ with vertex $\{1, k, N \}$,  as shown in Figure \ref{fig: interpolation_graphcial_model} (A-2). 
The potential between nodes 1 and $k$ is $\potl \defeq \psi^{(k-1)}$, and the potential between nodes $k$ and $N$ is $\potr \defeq \psi^{(N-k)}$, where $\psi^{(k)}$ is the $k$-th power of matrix $\psi \in \dR^{n \times n}$. 
From (\ref{formula: approx CGM}), the objective function of the approximated MAP inference can be calculated as
\begin{align}
  \mathcal{L}(\bc, \tpl, \tpr)
  &= \sum_{s \in \{1, 2\}}\sum_{i, j \in [n]} \left( T_{s ij}\log T_{s ij} - T_{s ij} \log \phi_{s ij} \right) - \sum_{i \in [n]} c_i \log c_i + \mathrm{const. }
\end{align}
and the feasible region is $\{\bc \in \dR_{\geq 0}^{n}, \tpl, \tpr \in \dR_{\geq 0}^{n \times n}\mid  \tpl \bone{n} = \ba, \tpl^\top \bone{n} = \bc, \tpr \bone{n} = \bc, \tpr^\top \bone{n} = \bb \}$, where $c_i \defeq z_2(i), T_{1ij} \defeq z_{1, k}(i, j), T_{2ij} \defeq z_{k, N}(i, j)$. 

This optimization problem can be solved by a message passing style algorithm for MAP inference of CGM \cite{Sun2015}\cite{Singh2020}. 
In this case, we can write the algorithm using only matrix multiplications, which resembles the Sinkhorn-Knopp algorithm. 
The algorithm is given in Algorithm \ref{alg: MP interpolation}.
Moreover, we can reduce the number of matrix multiplications by eliminating $\bx, \bz$ from the while loop in  Algorithm \ref{alg: MP interpolation} (shown in Algorithm \ref{alg: MP interpolation_simple}). 
This new algorithm has another merit: we can calculate interpolation results efficiently for all $k \in \{1, 
\ldots, N-2 \}$. 
In Algorithm \ref{alg: MP interpolation_simple}, lines 1--5 do not depend on $k$ because $\potl \potr = \psi^{(N-1)}$ is invariant for $k$. 
This makes it possible to precompute and reuse $\by$ and $\bz$ when calculating interpolation resulst for all $k$, leading to a significant reduction in computation burden. 

\newcommand{\Hs}{a}
\newcommand{\bHs}{\bm{\Hs}}
\newcommand{\Hm}{c}
\newcommand{\bHm}{\bm{\Hm}}
\newcommand{\He}{b}
\newcommand{\bHe}{\bm{\He}}
\begin{table}[t]
\begin{minipage}[t]{.48\linewidth}
  \small
  \begin{algorithm}[H]
    \DontPrintSemicolon
    \caption{Algorithm for histogram interpolation} \label{alg: MP interpolation}
    \KwIn{$\bHs, \bHe \in \dRp^n, \potl, \potr \in \dRp^{n \times n}$}
    \KwOut{result of interpolation $\bHm$}
    \BlankLine
    initialize: $\bx, \by, \bz, \bw \leftarrow \bone{n}$\;
    \While{not convergence}{
      $\bx \leftarrow \potl^\top \left( \bHs \oslash \bw \right)$\;
      $\by \leftarrow \potr^\top \bx$\;
      $\bx \leftarrow \potr \left( \bHe \oslash \by \right)$\;
      $\bw \leftarrow \potl \bz$\;
    }
    \KwRet{$\bx \odot \bz$}\;
    \tcp{$\odot$:element-wise multiplication}
    \tcp{$\oslash$:element-wise division}
  \end{algorithm}
\end{minipage}%
\hfill
\begin{minipage}[t]{.48\linewidth}
  \small
  \begin{algorithm}[H]
    \DontPrintSemicolon
    \caption{Simplified algorithm for histogram interpolation} \label{alg: MP interpolation_simple}
    \KwIn{$\bHs, \bHe \in \dRp^n, \potl, \potr \in \dRp^{n \times n}$}
    \KwOut{result of interpolation $\bHm$}
    \BlankLine
    initialize: $\by, \bw \leftarrow \bone{n}$\;
    \While{not convergence}{
      $\by \leftarrow \left( \potl \potr \right)^\top \left( \bHs \oslash \bw \right)$\;
      $\bw \leftarrow \left( \potl \potr \right) \left( \bHe \oslash \by \right)$\;
    }
    $\bx \leftarrow \potl^\top \left( \bHs \oslash \bw \right)$\;
    $\bz \leftarrow \potr \left( \bHe \oslash \by \right)$\;

    \KwRet{$\bx \odot \bz$}\;
    \tcp{$\odot$:element-wise multiplication}
    \tcp{$\oslash$:element-wise division}
  \end{algorithm}
\end{minipage}
\end{table}

\subsection{Continuous time Markov chain-based method} \label{subsec: continuous Markov}
The method proposed in \ref{subsec: undirected graphical model} has a disadvantage in that $t$ must be approximated as a rational number, and potential function $\psi$ has to be set according to the number of vertices of $N$.
We can avoid both issues by using a continuous time Markov \cite{Bremaud2013} chain as the underlying probabilistic model.

First, we briefly review the continuous time Markov chain. 
Let matrix $Q \in \dR^{n \times n}$ satisfy $Q \bone{n} = \bzero{n}$, $Q_{ii} < 0 \ (\forall \seq{i}{n})$, 
$Q_{ij} \geq 0 \ (\forall \seq{i, j}{n}), i \neq j)$. 
We consider the following process: (i) When the state changes to state $i\ (\seq{i}{n})$, the state stays in state $i$ for the duration drawn from an exponential distribution with mean $\frac{1}{-Q_{ii}}$. (ii) When the stay in state $i$ ends, the state transits to state $j$ with probability $\frac{Q_{ij}}{-Q_{ii}}$. 
This process is called continuous time Markov chain, and $Q$ is called transition rate matrix. 
Given initial distribution $\bpi_0$, the state distribution at time $t$ can be written as
$
  \bpi_t^\top = \bpi_0^\top \exp(t \cdot Q), 
$
where $\exp(\cdot)$ is matrix exponential function. 

In the proposed method, we consider a probabilistic model in which (i) the state at time $0$ is determined following the initial distribution $\bpi_0$ (ii) the state evolves until time $1$ according to the continuous time Markov chain with transition rate matrix $Q$. 
In this case, the transition probability matrix from time $0$ to $t$ is $\potl =\exp(t\cdot Q)$ and from time $t$ to $1$ is $\potr =\exp((1-t)\cdot Q)$. 
The corresponding graphical model is shown in Figure \ref{fig: interpolation_graphcial_model} (B). 
Even this directed graphical model allows us to construct a CGM and calculate the joint probability of contingency tables and observations using the probability mass function for a multinomial distribution. 
By utilizing this CGM, we can interpolate the histogram at arbitrary time $t$, in the same way as the method described in Section \ref{subsec: undirected graphical model}. 
This optimization problem is almost the same as the one described in Section \ref{subsec: undirected graphical model}, and can be solved via Algorithm \ref{alg: MP interpolation} or Algorithm \ref{alg: MP interpolation_simple} by replacing $\potl \leftarrow \exp(t\cdot Q)$ and $\potr \leftarrow \exp((1-t)\cdot Q)$. Please see Appendix for details. 
 
\subsection{Extension to interpolation on general trees}
The interpolation problem between two histograms can be generalized to interpolation problems on general graphs $G=(V, E)$, where we estimate histograms $(\ba_{u})_{u \in U}$ for some given set $U \subseteq V$ of nodes, given histograms $(\ba_{u})_{u \in V \setminus U}$ in the complementary set $V \setminus U$ of nodes. 
This framework can deal with various problems such as finding a barycenter between three or more histograms \cite{Solomon2015}. 
The interpolation problem discussed in the previous sections can be considered as a special case where the graph is a path graph and the histograms are observed at two leaves.
It is known that Wasserstein barycenter can be generalized to solve this kind of problems and this generalization is called Wasserstein Propagation \cite{Solomon2014}. 

The proposed method also can be generalized to address interpolation on general trees. 
To do this, we consider a CGM on the graphical model represented by tree $G = (V, E)$ and solve the MAP inference problem (\ref{formula: approx CGM}) under observations $(\ba_{u})_{u \in V \setminus U}$. The estimated node contingency table $\bz_{u}^*$ is the interpolated histogram on node ${u \in U}$. 
This MAP inference problem also can be efficiently solved by a message-passing type algorithm \cite{Sun2015}\cite{Singh2020}. Details of the algorithm are shown in Appendix.  

\section{Experimental results} \label{sec: experiments}
\begin{figure}[t]
  \begin{center}
  \includegraphics[width=140mm]{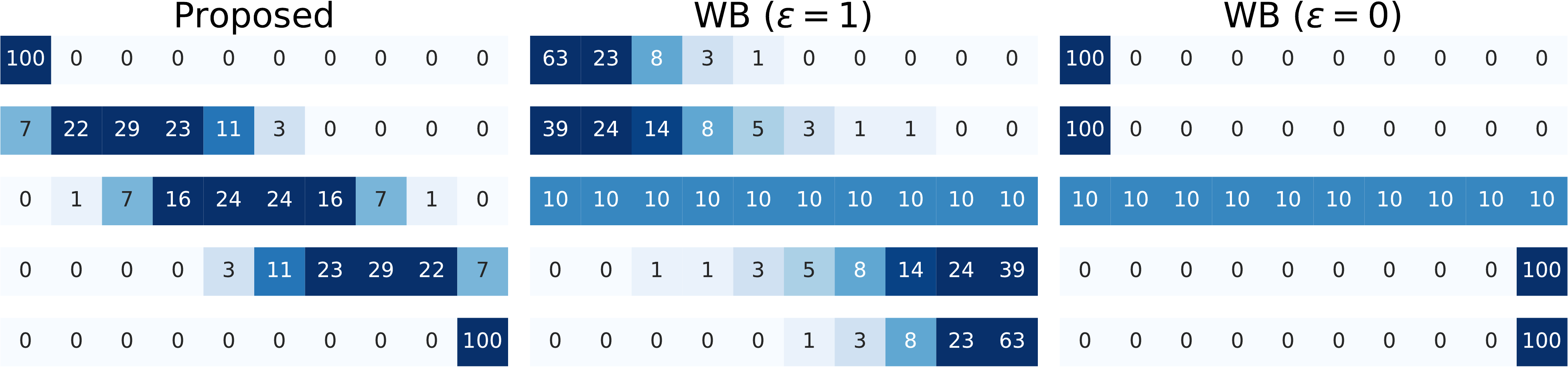}
  \end{center}
  \caption{Results of interpolation by two methods. For both methods, interpolation results for $t=$0, 0.25, 0.50, 0.75, 1 are placed at the top. Because there are infinitely many optimum solutions for WB ($\epsilon=0$) when $t=0.5$, we show one of them. Values in cells are rounded. } 
  \label{fig: line_comparison}
\end{figure}
\subsection{Synthetic data} \label{subsec: synthetic data}
We assumed a space with 10 cells arranged in a straight line, and considered interpolation between histograms $\ba$ and $\bb$ over this space. 
We set $\ba = (100, 0, \ldots, 0)$, $\bb = (0, \ldots, 0, 100)$. 
We compared the proposed method based on continuous time Markov chain (described in Section \ref{subsec: continuous Markov}) with Wasserstein Barycenter based on (\ref{formula: WB}) (WB in short) with $\epsilon = 0, 1.0$. 
For the proposed method, we used transition rate matrix $Q$ such that $Q_{ij}=1$ for adjacent cell pairs $(i, j)$, $Q_{ij}=0$ otherwise, and $Q_{ii} = - \sum_{j \neq i}Q_{ij}$.  
For WB, we set distances between adjacent cells to 1 and calculate the distance matrix between all cells; this matrix was used as cost matrix $C$. 
We used Algorithm \ref{alg: MP interpolation_simple} for the proposed method.  
For WB, we used analytical solutions, because analytical solutions can be calculated explicitly in this setting (for details, see the Appendix).

The results are shown in Figure \ref{fig: line_comparison}.
For all methods, interpolation results for $t=$0, 0.25, 0.50, 0.75, 1 are placed from the top. Because there are infinitely many optimum solutions for WB ($\epsilon=0$) when $t=0.5$, we show one of them. 
The result of WB ($\epsilon=0$) is the same as that with $\ba$ when $t \in [0, 0.5]$ and the same as that with $\bb$ when $t \in (0.5, 1.0)$;  there are infinitely many solutions when $t=0.5$. These results are not suitable in terms of interpolation. 
There are several differences between interpolation results of the proposed method and WB ($\epsilon=1$). 
First, the proposed method interpolates such that one flock moves as $t$ progresses, whereas the WB ($\epsilon=1$) interpolates such that the population spread out to all the cells. 
This characteristics of the proposed method is suitable for some applications, especially when we want to interpolate the way something is moving. 
Second, while the results of the proposed method for $t=0, 1$ are consistent with $\ba, \bb$, respectively, the results of WB ($\epsilon=1$) are not. 
This inconsistency is caused by the entropic regularization term. 
This property makes WB ($\epsilon=1$) hard to use, because interpolation results around given histograms are far from given histograms, leading to non-smooth interpolation.  

\subsection{Real data} \label{subsec: real data}
We evaluated the interpolation accuracy achieved with real-world spatio-temporal population data. 
We used mobile spatial statistics \cite{Terada2013}, which is the hourly population data for fixed square grids calculated from mobile network operation data. 
We used data in Tokyo and Kanagawa prefecture, which forms the main part of the capital area of Japan. 
The targeted area is divided into 2km $\times$ 2km square cells, 
and the data consist of population histograms of cells at $T$ -o'clock ($T \in\{0, \ldots, 23\}$) from April 1st, 2015 to April 30th, 2015. 
The number of cells $n$ is 196 and the total population in all the cells is about $1.8 \times 10^7$. 
$\bm{N}_{T, d}$ denotes the histogram of cell population at $T$ -o'clock on the $d$ -th day of the month. 
We calculated estimated population histogram $\hat{\bm{N}}_{T, d}$ from observed histograms at previous and next time, $\bm{N}_{T-1, d}$ and $\bm{N}_{T+1, d}$ using interpolation methods with $t=0.5$ ($(T \in \{1, \ldots, 22\}, d \in \{1, \ldots, 30\}$) and evaluated the discrepancy between $\hat{\bm{N}}_{T, d}$ and $\bm{N}_{T, d}$ by MAPE (Mean Absolute Percentage Error). 
For the proposed method, we used transition rate matrix $Q$ such that $Q_{ij}=q$ for adjacent cell pairs $(i, j)$, $Q_{ij}=0$ otherwise, and $Q_{ii} = - \sum_{j \neq i}Q_{ij}$.  
We calculated WB using POT: Python Optimal Transport library \cite{Flamary2017}.  The cost matrix $C$ for WB was given by the Euclidean distance between cells. 

\begin{figure}[t]
  \begin{center}
  \includegraphics[width=120mm]{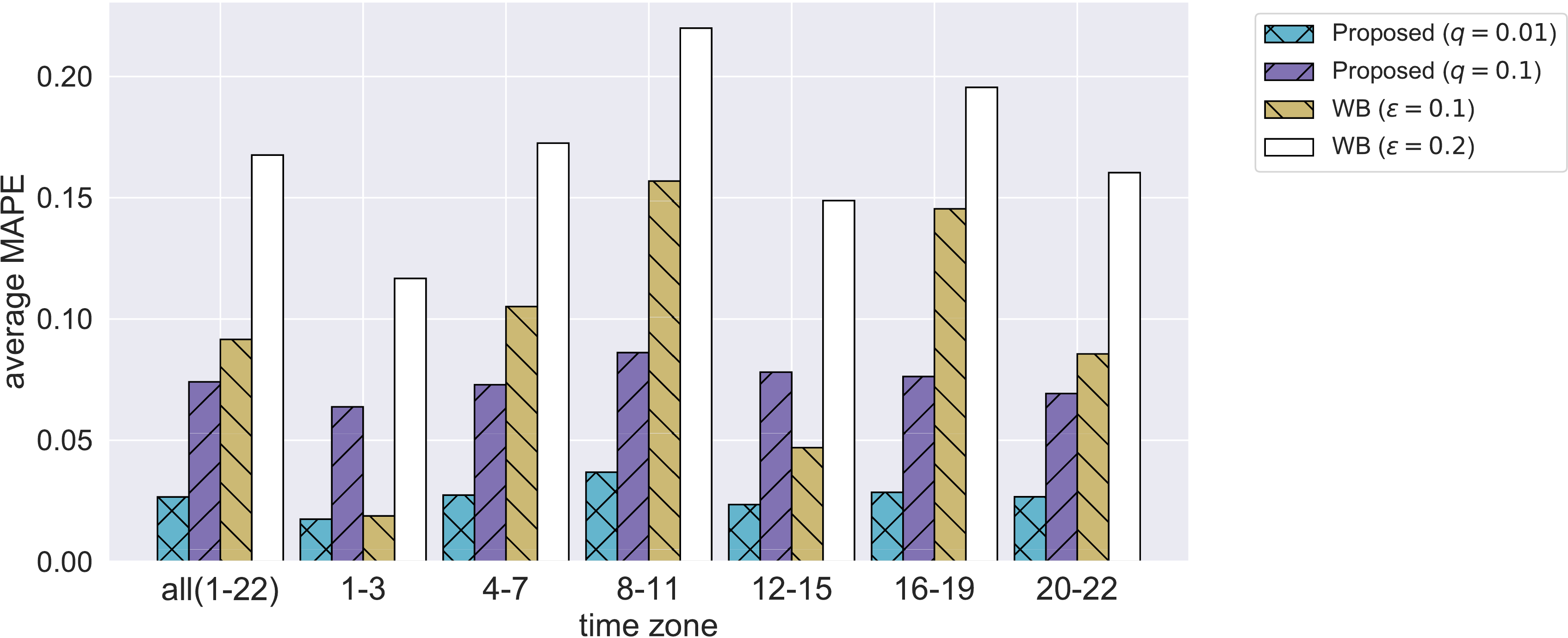}
  \end{center}
  \caption{Interpolation error of each method on population histogram data from Tokyo and Kanagawa prefectures. We used MAPE as the evaluation metric for evaluating the difference between interpolated histogram and true histogram. }
  \label{fig: real_error}
\end{figure}

Results are shown in Figure \ref{fig: real_error}. 
The leftmost bars are the 30-day averages of MAPE for all time zones, and the others are the 30-day averages of MAPE for each time zone (1--3, 4--7, 8--11, 12--15, 16--19, 20--22 o'clock, respectively). 
Performance improvements are attained by the proposed methods in total score and almost all time zones.
MAPEs of WB ($\epsilon=0.2$) are large in all time zones. This is caused by its large entropic regularization term, which blurs the estimation result excessively. 
MAPEs of WB ($\epsilon=0.1$) are small in time zones 1--3 and 12--15, but large in other time zones. 
This is caused by the difference in intensity of crowd movements in the targeted area. 
In time zones 1--3 and 12--15, the histogram shape does not change so much because people do not move around, but in other time zones the histograms change greatly because many people move around with commuting. 
WB ($\epsilon=0.1$) does not seem to be able to deal with the drastic histogram changes. 
We tried to calculate WB also for $\epsilon=0.01$, but the algorithm does not converge because $\epsilon$ is too small. 
The proposed methods achieve small MAPEs in all time zones, regardless of the intensity in crowd movement. 

\section{Conclusion}
This paper revealed the relationship between OT and CGM, and proposed a new framework in which OT is interpreted as a MAP solution of a CGM. Based on this insight, we proposed OT with noisy observations and a new interpolation method between histograms.   
Experiments showed the effectiveness of the proposed interpolation method. 
It will be interesting future work to apply our probabilistic approach to other OT-related tasks, such as ground metric learning \cite{Cuturi2014gml}. 

\bibliographystyle{plain}
\bibliography{main}

\begin{thebibliography}{10}

\bibitem{Agueh2011}
Martial Agueh and Guillaume Carlier.
\newblock Barycenters in the {{W}}asserstein space.
\newblock {\em SIAM Journal on Mathematical Analysis}, 43(2):904--924, 2011.

\bibitem{Arjovsky2017}
Martin Arjovsky, Soumith Chintala, and L{\'e}on Bottou.
\newblock Wasserstein generative adversarial networks.
\newblock In {\em ICML}, pages 214--223, 2017.

\bibitem{Benamou2003}
Jean-David Benamou.
\newblock Numerical resolution of an “unbalanced” mass transport problem.
\newblock {\em ESAIM: Mathematical Modelling and Numerical Analysis},
  37(5):851--868, 2003.

\bibitem{Blondel2018}
Mathieu Blondel, Vivien Seguy, and Antoine Rolet.
\newblock Smooth and sparse optimal transport.
\newblock In {\em AISTATS}, volume~84 of {\em Proceedings of Machine Learning
  Research}, pages 880--889. {PMLR}, 2018.

\bibitem{Bremaud2013}
Pierre Br{\'e}maud.
\newblock {\em {{M}}arkov chains: {{G}}ibbs fields, {{M}}onte {{C}}arlo
  simulation, and queues}, volume~31.
\newblock Springer Science \& Business Media, 2013.

\bibitem{Chizat2018}
Lena{\"{\i}}c Chizat, Gabriel Peyr{\'{e}}, Bernhard Schmitzer, and
  Fran{\c{c}}ois{-}Xavier Vialard.
\newblock Scaling algorithms for unbalanced optimal transport problems.
\newblock {\em Mathematics of Computation}, 87(314):2563--2609, 2018.

\bibitem{Courty2017}
Nicolas Courty, R{\'e}mi Flamary, Devis Tuia, and Alain Rakotomamonjy.
\newblock Optimal transport for domain adaptation.
\newblock {\em IEEE Transactions on Pattern Analysis and Machine Intelligence},
  39(9):1853--1865, 2017.

\bibitem{Cuturi2013}
Marco Cuturi.
\newblock Sinkhorn distances: Lightspeed computation of optimal transport.
\newblock In {\em NIPS}, pages 2292--2300, 2013.

\bibitem{Cuturi2014gml}
Marco Cuturi and David Avis.
\newblock Ground metric learning.
\newblock {\em The Journal of Machine Learning Research}, 15(1):533--564, 2014.

\bibitem{Flamary2017}
R{'e}mi Flamary and Nicolas Courty.
\newblock Pot python optimal transport library, 2017.

\bibitem{Frogner2015}
Charlie Frogner, Chiyuan Zhang, Hossein Mobahi, Mauricio Araya, and Tomaso~A
  Poggio.
\newblock Learning with a {{W}}asserstein loss.
\newblock In {\em NIPS}, pages 2053--2061, 2015.

\bibitem{Knight2008}
Philip~A Knight.
\newblock The {{S}}inkhorn--{{K}}nopp algorithm: convergence and applications.
\newblock {\em SIAM Journal on Matrix Analysis and Applications},
  30(1):261--275, 2008.

\bibitem{Nguyen2016}
Thien Nguyen, Akshat Kumar, Hoong~Chuin Lau, and Daniel Sheldon.
\newblock Approximate inference using {{DC}} programming for collective
  graphical models.
\newblock In {\em AISTATS}, pages 685--693, 2016.

\bibitem{Peyre2019}
Gabriel Peyr{\'e} and Marco Cuturi.
\newblock Computational optimal transport.
\newblock {\em Foundations and Trends{\textregistered} in Machine Learning},
  11(5-6):355--607, 2019.

\bibitem{Roberts2017}
Lucas Roberts, Leo Razoumov, Lin Su, and Yuyang Wang.
\newblock {{G}}ini-regularized optimal transport with an application to
  spatio-temporal forecasting.
\newblock 2017.
\newblock arXiv:1712.02512.

\bibitem{Sheldon2013a}
Daniel Sheldon, Tao Sun, Akshat Kumar, and Tom Dietterich.
\newblock Approximate inference in collective graphical models.
\newblock In {\em ICML}, pages 1004--1012, 2013.

\bibitem{Sheldon2011}
Daniel~R. Sheldon and Thomas~G. Dietterich.
\newblock Collective graphical models.
\newblock In {\em NIPS}, pages 1161--1169, 2011.

\bibitem{Singh2020}
Rahul Singh, Isabel Haasler, Qinsheng Zhang, Johan Karlsson, and Yongxin Chen.
\newblock Inference with aggregate data: An optimal transport approach.
\newblock 2020.
\newblock arXiv:2003.13933.

\bibitem{Solomon2015}
Justin Solomon, Fernando De~Goes, Gabriel Peyr{\'e}, Marco Cuturi, Adrian
  Butscher, Andy Nguyen, Tao Du, and Leonidas Guibas.
\newblock Convolutional {{W}}asserstein distances: Efficient optimal
  transportation on geometric domains.
\newblock {\em ACM Transactions on Graphics}, 34(4):1--11, 2015.

\bibitem{Solomon2014}
Justin Solomon, Raif Rustamov, Leonidas Guibas, and Adrian Butscher.
\newblock {{W}}asserstein propagation for semi-supervised learning.
\newblock In {\em ICML}, pages 306--314, 2014.

\bibitem{Sun2015}
Tao Sun, Daniel Sheldon, and Akshat Kumar.
\newblock Message passing for collective graphical models.
\newblock In {\em ICML}, pages 853--861, 2015.

\bibitem{Terada2013}
Masayuki Terada, Tomohiro Nagata, and Motonari Kobayashi.
\newblock Population estimation technology for mobile spatial statistics.
\newblock {\em NTT DOCOMO Technical Journal}, 14(3):10--15, 2013.

\end{thebibliography}


\begin{thebibliography}{}
  \bibitem[23]{Moler2003}
  Cleve Moler, Charles Van Loan. 
 \newblock {Nineteen dubious ways to compute the exponential of a matrix, twenty-five years later}.
 \newblock {\em SIAM review}, 
 \newblock 45(1):3--49, 
 \newblock 2003. 
\end{thebibliography}

\newpage
\appendix
\begin{center}
 \large\textbf{Supplementary Material: Probabilistic Optimal Transport based on Collective Graphical Models}
\end{center}
\section{Proof of Proposition \ref{prop: objective}}
\begin{proof}
  Because 
  \begin{align*}
    \mathcal{L}(T) 
    &= \sum_{i, j \in [n]} \left[ (- \log \psi_{ij}) \cdot  \TP_{ij} + \TP_{ij} \log \TP_{ij} \right] - F \log F + F \log Z  \quad (\because (\ref{formula: approx CGM}))  \\
    &= \sum_{i, j \in [n]} \left[ (- \log \psi_{ij}) \cdot  F \tau_{ij} + F \tau_{ij} \log F  + F \tau_{ij} \log \tau_{ij} \right] - F \log F + F \log Z \\
    &= F \cdot \sum_{i, j \in [n]} \left[ C_{ij} \cdot \tau_{ij} + \tau_{ij} \log \tau_{ij} \right] + F \log Z \quad (\because \sum_{i, j \in [n]} \tau_{ij} = 1 ) \\
    &= F \cdot \left( \mathcal{G}_C^1(\tau) + \log Z \right), 
  \end{align*}
  we have $\mathcal{G}_C^1(\tau) = (1/F) \cdot \mathcal{L}(T)  - \log Z$. 
\end{proof}
\section{Proof of Proposition \ref{proposition: OT vs CGM}}
\begin{proof} 
  From Proposition \ref{prop: objective}, we have
  \begin{align*}
    \mathcal{D}_C^1(\bm{\alpha}, \bm{\beta})
    = \min_{\tau \in \UU{\bm{\alpha}}{\bm{\beta}}} \mathcal{G}_C^1(\tau) 
    = \min_{T \in \UU{\ba}{\bb}} \left[ \frac{1}{F} \cdot  \mathcal{L}(T) - \log Z \right]
    = \frac{1}{F} \cdot  \min_{T \in \UU{\ba}{\bb}} \left[  \mathcal{L}(T) \right] - \log Z.
  \end{align*}
\end{proof}

\section{Derivation of OT with noisy observation}
\subsection{Gaussian noise case}
\begin{align*}
  &\mathcal{G}_C^1(\tau) - \frac{1}{F} \log \Pr(F \bm{\alpha} \mid F \tau \bone{n}) - \frac{1}{F} \log \Pr(F \bm{\beta} \mid F \tau^\top \bone{n}) \\
  &= \mathcal{G}_C^1(\tau)
  + \frac{1}{F} \sum_{i \in [n]} \left[ \frac{\left\{ F \alpha_i - F (\tau \bone{n})_i \right\}^2 }{2 F \sigma^2} \right] 
  + \frac{1}{F} \sum_{i \in [n]} \left[ \frac{\left\{ F \beta_i - F (\tau^\top \bone{n})_i \right\}^2 }{2 F \sigma^2} \right]
  + \frac{n \log(2 \pi F \sigma^2)}{F} \\
  &= \mathcal{G}_C^1(\tau)
  + \frac{1}{2 \sigma^2} \sum_{i \in [n]} \left( \alpha_i - (\tau \bone{n})_i \right)^2
  + \frac{1}{2 \sigma^2} \sum_{i \in [n]} \left( \beta_i - (\tau^\top \bone{n})_i \right)^2
  + \frac{n \log(2 \pi F \sigma^2)}{F} \\
  &\to \mathcal{G}_C^1(\tau) + \| \bm{\alpha} - \tau \bone{n} \|_2^2 + \| \bm{\beta} - \tau^\top \bone{n} \|_2^2 \quad (F \to \infty). 
\end{align*}

\subsection{Poisson noise case}
Because 
\begin{align*}
  - \log \left( \frac{y^x e^{-y}}{x!} \right)
  &= -x \log y + y + \log x! \\
  &\approx  -x \log y + y + x \log x - x \quad (\because \mathrm{Stirling's\ approximation}) \\
  & = x \log \frac{x}{y} - x + y,  
\end{align*}
we have
\begin{align*}
  &\mathcal{G}_C^1(\tau) - \frac{1}{F} \log \Pr(F \bm{\alpha} \mid F \tau \bone{n}) - \frac{1}{F} \log \Pr(F \bm{\beta} \mid F \tau^\top \bone{n}) \\
  &= \mathcal{G}_C^1(\tau)
  + \frac{1}{F} \sum_{i \in [n]} 
  \left[ 
  F \alpha_i \log \frac{F \alpha_i}{(F \tau \bone{n})_i} - F \alpha_i + (F \tau \bone{n})_i
  \right] \\
  &\quad + \frac{1}{F} \sum_{i \in [n]} 
  \left[ 
  F \beta_i \log \frac{F \beta_i}{(F \tau^\top \bone{n})_i} - F \beta_i + (F \tau^\top \bone{n})_i
  \right] \\
  &= \mathcal{G}_C^1(\tau)
  + \sum_{i \in [n]} 
  \left[ 
  \alpha_i \log \frac{\alpha_i}{(\tau \bone{n})_i} - \alpha_i + (\tau \bone{n})_i
  \right] 
  + \sum_{i \in [n]} 
  \left[ 
  \beta_i \log \frac{\beta_i}{(\tau^\top \bone{n})_i} -  \beta_i + ( \tau^\top \bone{n})_i
  \right] \\
  &= \mathcal{G}_C^1(\tau) 
  + \gKL(\bm{\alpha} \| \tau \bone{n})
  + \gKL(\bm{\beta} \| \tau^\top \bone{n}). 
\end{align*}

\section{Optimization algorithm of OT with noisy observation}
Let $A: \dR^n \to \dR \cup \{+\infty\}$ and $B: \dR^n \to \dR \cup \{+\infty\}$ are convex functions.
It is known that minimization problem
\begin{align}
  \min_{\tau \in \dR_{\geq 0}^{n \times n}} \left[ \mathcal{G}_C^1(\tau) + A(\tau \bone{n}) + B(\tau^\top \bone{n})\right], \quad
  \mathcal{G}_C^1(\tau) = \sum_{i \in [n]} \sum_{j \in [n]} \left( -\log \psi_{ij} \tau_{ij} + \tau_{ij} \log \tau_{ij} \right)
\end{align}
can be solved by scaling algorithm described by following iterations \cite{Chizat2018}:
\begin{align}
  \bu \leftarrow \KLprox{A}{\psi \  \bv} \oslash \left( \psi \  \bv \right),
  \bv \leftarrow \KLprox{B}{\psi^\top \ \bu} \oslash \left( \psi^\top \  \bu \right), 
\end{align}
where $\oslash$ is element-wise division and $\KLprox{A}{\cdot}$ is the proximal operator for KL divergence: 
\begin{align}
  \forall \bu \in \mathbb{R}_{\geq 0}^n, \  \KLprox{A}{\bu} \defeq \argmin_{\bu' \in \mathbb{R}_{\geq 0}^n} \left[ \KL(\bu' \mid \bu) + A(\bu') \right].  
\end{align}

Thus, for noise distribution $P_1(\ba, \bta) \defeq \Pr(\ba \mid \bta)$ and $P_2(\bb, \btb) \defeq \Pr(\bb \mid \btb)$, 
by setting $A(\bx) = -(1/F) \log P_1(F \bm{\alpha} \mid F \bx) $ and $B(\bx) = -(1/F) \log P_2(F \bm{\beta} \mid F \bx)$, we can solve the optimization problem (\ref{formula: noisy OT}) via scaling algorithm described above. For more details, please see \cite{Chizat2018}.

\section{Computational complexity of Algorithm \ref{alg: MP interpolation} and Algorithm \ref{alg: MP interpolation_simple}}
\subsection{Undirected graphical model-based method}
If we implement these algorithms naively, precomputation of $\phi_1 = \psi^{(k-1)}$ and $\phi_2 = \psi^{(N-k)}$ takes $O(n^3 \log N)$ time by square-and-multiply algorithm, and matrix calculation in the while loop takes $O(n^2)$ time per one loop. 
Thus, the total time complexity is $O(n^3 + J n^2)$, where $J$ is the number of iterations. 

There is another method; 
in Algorithm \ref{alg: MP interpolation} and Algorithm \ref{alg: MP interpolation_simple}, we don't need the matrices $\phi_1$, $\phi_2$ and $\phi_1 \phi_2$ but only the products $\phi_1 \bx$, $\phi_2 \bx$ and $\phi_1 \phi_2 \bx$ for some vector $\bx$. 
We can calculate these products in $O(\mathrm{nnz}(\psi) \cdot N)$ time, where $\mathrm{nnz}(\psi)$ is the number of  non-zero elements of $\psi$, by calculating $\psi \bx, \psi (\psi \bx), \psi (\psi (\psi \bx)), \ldots$ in order.  
Using this method, precomputation is not needed and matrix calculation in while loop takes $O(\mathrm{nnz}(\psi) \cdot N)$ time per one loop, so the total time complexity is $O(J \cdot \mathrm{nnz}(\psi) \cdot N)$. 
When potential matrix $\psi$ is sparse and $N$ is small, this method is significantly efficient compare to the naive method. 

\subsection{Continuous time Markov chain-based method}
As in the case of undirected graphical model-based method, we only need $\exp({t \cdot Q}) \bx$, $\exp((1-t)\cdot Q) \bx$, $\exp(Q) \bx$ for some vector $\bx$. 
These values van be calculated in  $O(\mathrm{nnz}(Q))$ time \cite{Moler2003}. 
Therefore, matrix calculation in while loop takes $O(\mathrm{nnz}(Q))$ time per one loop and the total time complexity is $O(J \cdot \mathrm{nnz}(Q))$. 

\section{Derivation of MAP inference problem of the directed CGM}
From the generation process of samples in the probabilistic model described in Section \ref{subsec: continuous Markov}, the joint probability can be calculated as follows:
\begin{align*}
  &\Pr(\ba, \bb, \bc, T_1, T_2 ; \bpi_0, \phi_1, \phi_2)
  = \Pr(\ba ; \bpi_0) \Pr(T_1 \mid \ba ; \phi_1) \Pr(T_2 \mid \bc;\phi_2) \\
  &= \frac{F!}{\prod_{i \in [n]} a_i!} \prod_{i \in [n]} \pi_{0i}^{a_i} \cdot 
  \prod_{i \in [n]} \left( \frac{a_i!}{\prod_{j \in [n]} T_{1ij}!} \prod_{j \in [n]} \phi_{1ij}^{T_{1ij}} \right) \cdot
  \prod_{i \in [n]} \left( \frac{c_i!}{\prod_{j \in [n]} T_{2ij}!} \prod_{j \in [n]} \phi_{2ij}^{T_{2ij}} \right)
\end{align*}
if $\tpl \bone{n} = \ba, \tpl^\top \bone{n} = \bc, \tpr \bone{n} = \bc, \tpr^\top \bone{n} = \bb$, and  
$
  \Pr(\ba, \bb, \bc, T_1, T_2 ; \bpi_0, \phi_1, \phi_2) = 0
$
otherwise. 
Thus, for $(\bc, T_1, T_2)$ which satisfy $\tpl \bone{n} = \ba, \tpl^\top \bone{n} = \bc, \tpr \bone{n} = \bc, \tpr^\top \bone{n} = \bb$, 
\begin{align*}
  &- \log \Pr(\ba, \bb, \bc, T_1, T_2 ; \bpi_0, \phi_1, \phi_2) \\
  &= \sum_{s \in \{1, 2\}}\sum_{i, j \in [n]} \left( \log T_{s ij}! - T_{s ij} \log \phi_{s ij} \right) - \sum_{i \in [n]} \log c_i ! - \log F! -\sum_{i \in [n]} a_i \log \pi_{0i}\\
  &\approx \sum_{s \in \{1, 2\}}\sum_{i, j \in [n]} \left( T_{s ij} \log T_{s ij} - T_{s ij} - T_{s ij} \log \phi_{s ij} \right) - \sum_{i \in [n]} (c_i \log c_i -c_i) \\
  &- (F \log F - F) -\sum_{i \in [n]} a_i \log \pi_{0i} \quad (\because \mathrm{Stirling's \ approximation})\\
  &= \sum_{s \in \{1, 2\}}\sum_{i, j \in [n]} \left( T_{s ij}\log T_{s ij} - T_{s ij} \log \phi_{s ij} \right) - \sum_{i \in [n]} c_i \log c_i - F \log F -\sum_{i \in [n]} a_i \log \pi_{0i} \\
  &= \sum_{s \in \{1, 2\}}\sum_{i, j \in [n]} \left( T_{s ij}\log T_{s ij} - T_{s ij} \log \phi_{s ij} \right) - \sum_{i \in [n]} c_i \log c_i + \mathrm{const. }, 
\end{align*}
where we used $\sum_{i \in [n]} \sum_{j \in [n]} T_{sij} = F$ for $s \in \{ 1, 2\}$  and $\sum_{i \in [n]} c_i = F$. 

\section{Interpolation algorithm on general trees}
For tree $G = (V, E)$ and given histograms $(\ba_{u} \in \Sigma_n^F)_{u \in V \setminus U}$, we consider an optimization problem below:
\footnotesize
\begin{align} \label{opt: tree}
  \begin{aligned}
  \min_{\bz} & && \sum_{(u, v) \in E} \sum_{i, j \in [n]} z_{uv}(i, j) \left( \log z_{uv}(i, j) -  \log \phi_{uv}(i, j) \right) -  \sum_{u \in V} (\nu_u-1)\sum_{i \in [n]} z_u(i) \log z_u(i),  \\
  \mathrm{s.t.} &&& F = \sum_{i \in [n]} z_{u}(i) \quad \forall u \in V,  \\
  &&&  z_{u} (i) = \sum_{j \in [n]} z_{uv}(i, j) \quad \forall (u, v) \in E, \ \forall i \in [n], \\
  & && z_{u}(i) = a_{ui} \quad \forall i \in V \setminus U,\  \forall i \in [n]. \\
  \end{aligned}
\end{align}
\normalsize
The interpolated histogram on node $v \in U$ is given by $\bz^*_u$, where $\bz^*$ is the optimum solution of (\ref{opt: tree}). 

We can solve this optimization problem by message passing style algorithm, which is called Sinkhorn Belief Propagation in \cite{Singh2020}. For more details, please see \cite{Singh2020}. 

\section{Analytical solution of WB in synthetic data experiment}
We assume a space with $n$ cells arranged in a straight line, and consider interpolation between histograms $\ba = (F, 0, \ldots, 0)$ and $\bb = (0, \ldots, 0, F)$ over this space. 
The cost function (distance) between cell $i$ and $j$ is given by $C_{ij} = |i-j|$. 

Because
\begin{align}
  \mathcal{D}^{\epsilon}_{C}(\ba, \bc) &= \sum_{i \in [n]} \left( (i-1) \cdot c_i + \epsilon \cdot c_i \log c_i \right), \\ 
  \mathcal{D}^{\epsilon}_{C}(\bb, \bc) &= \sum_{i \in [n]} \left( (n-i) \cdot c_i + \epsilon \cdot c_i \log c_i \right), 
\end{align}
we have
\begin{align}
   (1-t) \cdot \mathcal{D}^{\epsilon}_{C}(\ba, \bc) + t \cdot \mathcal{D}^{\epsilon}_{C}(\bb, \bc) 
   &= \sum_{i \in [n]} \left( k_i c_i + \epsilon \cdot c_i \log c_i \right), \label{formula: WB objective}  
\end{align}
where $k_i  \defeq (1-2 t) \cdot i + tn + t - 1$. 
All we have to do is minimize this function under constraints $\sum_{i \in [n]} c_i = F, c_i \geq 0 \ (\forall i \in  [n])$. 

When $\epsilon = 0$, the objective function is $\sum_{i \in [n]}  k_i c_i $. 
For $t=1/2$, arbitrary $\bc \in \Sigma_{n}^F$ is optimum because $k_i$ take the same value for all $i \in [n]$. 
For $t<1/2$, optimum solution is $\ba$ since $k_1 < k_i$ for $i \in \{2, \ldots, n\}$. 
For $t>1/2$, optimum solution is $\bb$ since $k_n < k_i$ for $i \in \{1, \ldots, n-1\}$. 

When $\epsilon > 0$, let $L(\bc, \lambda)$ be the Lagrangian of Equation (\ref{formula: WB objective}) for the equality constraint $\sum_{i \in [n]} c_i = F$: 
\begin{align}
  L(\bc, \lambda) = \sum_{i \in [n]} \left( k_i c_i + \epsilon \cdot c_i \log c_i \right) 
  + \lambda \left( \sum_{i \in [n]} c_i - F \right). 
\end{align}
For all $i \in [n]$, $\partial L / \partial c_i = 0 \Rightarrow c_i = \Lambda e^{- k_i / \epsilon}$, where $\Lambda \defeq e^{- \left( \lambda/ \epsilon \right) - 1}$. 
Since $ \sum_{i \in [n]} c_i = F$, we get
\begin{align}
  \Lambda = \frac{F}{ \sum_{i \in [n]} e^{- k_i / \epsilon}},\quad c_i =  \frac{F}{ \sum_{i \in [n]} e^{- k_i / \epsilon} } \cdot  e^{- k_i / \epsilon}. 
\end{align}

\section{Details of real data experiments}
We here show details of real data experiments. 
We used Python3 to implement the algorithms, and
we conducted all experiments on a 64-bit macOS machine with Intel Core i7 CPUs and 16 GB RAM.
The results presented in Table \ref{table: real_error} are the same as written in Section \ref{subsec: real data} in the body except that standard deviations are written. 
Each result in Table \ref{table: real_error} is average and standard deviation of MAPE (Mean Absolute Percentage Error) in each time zone of 30 days (from April 1st, 2015 to April 30th, 2015). Standard deviation is shown in parentheses.
MAPE at $T$-o'clock on the $d$-th day is calculated by 
\begin{align}
  \frac{1}{n} \sum_{i \in [n]} \left| \frac{N_{T, d, i} - \hat{N}_{T, d, i}}{N_{T, d, i}} \right|, 
\end{align}
where $\bm{N}_{T, d}$ is the true histogram and $\hat{\bm{N}}_{T, d}$ is the estimated histogram. 
\begin{table*}[h] 
  \caption{The average and standard deviation of MAPE (Mean Absolute Percentage Error). The best result is highlighted for each time zone. Standard deviation is shown in parentheses. } \label{table: real_error}
  \begin{tabular}{|l|cccc|} \hline
    & \multicolumn{4}{|c|}{Time zones} \\ \cline{2-5}
    & All (1--23) & 1--3 & 4--7 & 8--11 \\ \hline
    Proposed ($q = 0.01$) & $\bm{0.027}$ (0.011) & $\bm{0.017}$ (0.003) & $\bm{0.027}$ (0.015) & $\bm{0.037}$ (0.013) \\
    Proposed ($q = 0.1$) & 0.075 (0.012) & 0.064 (0.005) & 0.073 (0.014) & 0.086 (0.013) \\
    WS ($\epsilon = 0.1$) & 0.097 (0.095) & 0.019 (0.004) & 0.105 (0.126) & 0.157 (0.120) \\
    WS ($\epsilon = 0.2$) & 0.172 (0.068) & 0.117 (0.007) & 0.173 (0.093) & 0.220 (0.091) \\ \hline
  \end{tabular}
  \\
  \\
  \\
  \begin{tabular}{|l|ccc|} \hline
    & \multicolumn{3}{|c|}{Time zones} \\ \cline{2-4}
    & 12--15 & 16--19 & 20-22 \\ \hline
    Proposed ($q = 0.01$) & $\bm{0.023}$ (0.003) & $\bm{0.029}$ (0.007) & $\bm{0.027}$ (0.005) \\
    Proposed ($q = 0.1$) & 0.078 (0.005) & 0.076 (0.007) & 0.069 (0.005) \\
    WS ($\epsilon = 0.1$) & 0.047 (0.021) & 0.145 (0.072) & 0.086 (0.038) \\
    WS ($\epsilon = 0.2$) & 0.149 (0.009) & 0.196 (0.044) & 0.160 (0.028)
    \\ \hline
  \end{tabular}
\end{table*}

\end{document}